\documentclass[lettersize,journal]{IEEEtran}
\usepackage{amsmath,amsfonts}
\usepackage{algorithmic}
\usepackage{algorithm}
\usepackage{array}
\usepackage[caption=false,font=normalsize,labelfont=sf,textfont=sf]{subfig}
\usepackage{textcomp}
\usepackage{stfloats}
\usepackage{url}
\usepackage{verbatim}
\usepackage{graphicx}
\usepackage{cite}
\usepackage[hidelinks]{hyperref}
\usepackage{multirow}
\usepackage{tabularx}
\usepackage{tabularray}
\usepackage{makecell}
\usepackage{xcolor}
\usepackage{booktabs}
\usepackage{pifont}
\usepackage{tikz}

\newif\ifjournal 
\journalfalse

\begin{document}

\title{Learning Unified Video and Image Representation for Video Face Forgery Detection}

\ifjournal
\else
\author{Haotian Liu, Yang Liu, Guoying Zhao, and Xiaobai Li}
\fi

\markboth{Journal of \LaTeX\ Class Files,~Vol.~14, No.~8, August~2021}%
{Shell \MakeLowercase{\textit{et al.}}: A Sample Article Using IEEEtran.cls for IEEE Journals}

\IEEEpubid{0000-0000/00\$00.00~\copyright~2021 IEEE}

\maketitle

\begin{abstract}
Face forgery detection is crucial for preserving the security and integrity of facial data given the rapid developments in face manipulation techniques and deep generative models. Existing methods for video face forgery detection typically assume that all frames in a forged video are manipulated, while detecting partially forged videos that contain only a subset of altered frames remains challenging. To address this issue, we propose a novel framework, UVIF, that utilizes additional annotated images to provide fine-grained supervision for detecting partial forgeries in videos. UVIF employs a unified encoder and a multi-task learning paradigm to jointly model facial videos and images for boosted video face forgery detection. A 2D backbone with temporal fusion modules is employed as the unified encoder. A pseudo labeling process is designed for video frames to bridge their representations with those of static images. A video-oriented feature alignment strategy is further introduced to reduce the distribution gap between videos and images. Extensive experiments on benchmark datasets demonstrate the effectiveness of our framework, which outperforms state-of-the-art methods in detecting partially forged videos while introducing no additional computational overhead.
Our code is available at \url{https://github.com/haotianll/UVIF}.
\end{abstract}


\section{Introduction}
\label{section:intro}
Face forgery detection aims to distinguish between authentic and fabricated faces \cite{survey_ijcv_juefei2022}.
This task is essential in preventing malicious uses of face manipulation techniques and AI-generated content (AIGC) \cite{GAN_goodfellow2020, stable_diffusion_rombach2022}, thereby upholding the security and integrity of facial data.

Current research on video forgery detection is primarily categorized into image-based and video-based methods.
Image-based methods \cite{mesonet_afchar2018,faceforensics++_rossler2019,madd_zhao2021,frequency_qian2020,self_blended_shiohara2022,recce_cao2022,hierarchical_liu2024,narrowing_yu2024,effort_yan2025orthogonal,deepfake_xu2026}
utilize static facial images as inputs and perform image classification to verify their authenticity. When dealing with a video clip, these methods extract individual facial frames and assign classification labels to each frame. 
In contrast, video-based methods \cite{ictu_li2018,lips_haliassos2021,biological_ciftci2020,stil_gu2021, forgerynet_he2021, ftcn_zheng2021,  altfreezing_wang2023, tall_xu2023, mintime_coccomini2024, pwtf_kim2025,dip_nie2025, dfd_fcg_han2025} process facial video clips directly using 3D backbones and perform video classification. Given the importance of temporal inconsistencies across video frames for forgery analysis, video-based methods \cite{ictu_li2018,lips_haliassos2021,biological_ciftci2020,stil_gu2021, forgerynet_he2021, ftcn_zheng2021,  altfreezing_wang2023, tall_xu2023, mintime_coccomini2024, pwtf_kim2025,dip_nie2025, dfd_fcg_han2025} achieved higher accuracy than image-based methods.

\begin{figure}[!t]
    \centering
    \includegraphics[width=\linewidth]{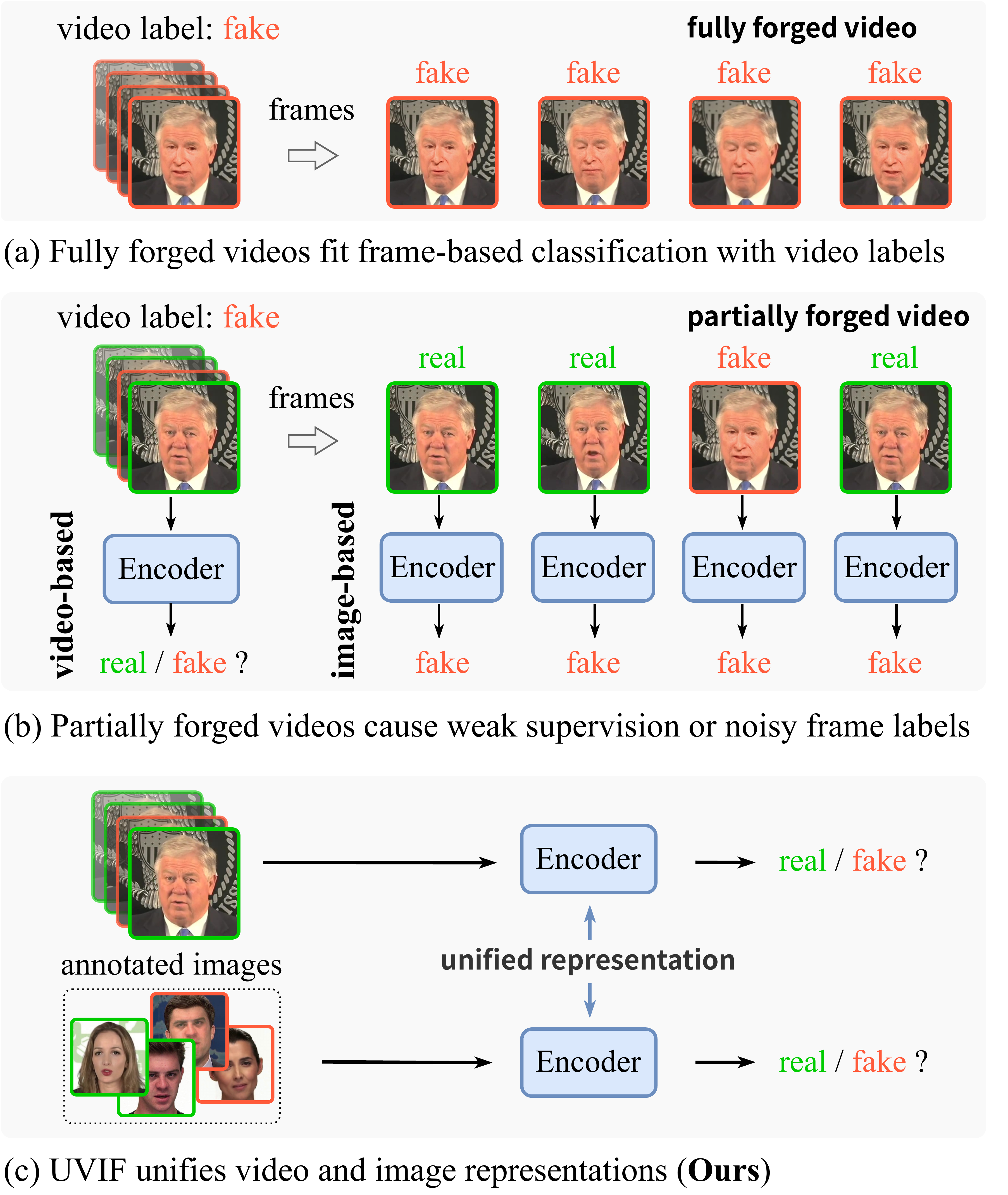}
    \vspace{-2em}
    \caption{(a) Fully forged video, the main focus of previous methods, where every frame in a fake video is manipulated and the video label can be assigned to each frame. (b) Partially forged video, where only a subset of frames is manipulated. Video-based methods lack frame-level supervision, while image-based methods assign the fake label to all frames, including real ones, introducing label noise during classification training. (c) The proposed UVIF uses a set of annotated images to provide fine-grained supervision for videos and learns unified representations.}
    \label{figure:motivation}
    \vspace{-1em}
\end{figure}

A significant limitation of existing face forgery detection studies is the assumption \cite{faceforensics++_rossler2019} that all frames in a fake video are manipulated, which underlies the preprocessing pipelines, model architectures, and training strategies employed.
However, this assumption does not hold for all face forgery detection tasks in realistic scenarios, as some fake videos may contain only a subset of manipulated frames. This discrepancy poses challenges for both image-based and video-based detection methods, as illustrated in Figure~\ref{figure:motivation}.
Video-based methods lack frame-level supervision and therefore require tailored strategies to handle mixtures of authentic and forged frames. 
Image-based methods instead assign the fake video label to every extracted frame, including authentic ones, which introduces label noise during image classification training.

\ifjournal
\else
\IEEEpubidadjcol
\fi

Some previous work \cite{mil_sharp_li2020} tried to use multiple instance learning (MIL) \cite{miml_deep_feng2017,mil_attention_ilse2018} to detect partially forged videos.
Specifically, MIL uses a set of labeled bags containing multiple instances for training. For binary classification, a positive bag may contain both positive and negative instances, while a negative bag contains only negative ones.
MIL matches the setting of partially forged video detection, where a forged video can be treated as a positive bag.
However, current MIL methods \cite{miml_deep_feng2017,mil_sharp_li2020,mil_attention_ilse2018,mil_dual_li2021,transmil_shao2021,acmil_zhang2024} group instances based on feature similarity without access to instance-level labels during training, limiting their effectiveness for partially forged video detection.

This limitation motivates the use of frame-level supervision, as detecting forged frames is essential for verifying the authenticity of facial videos.
Knowledge from detecting forged images can also contribute to this, as they have a similar representation learning process, i.e., extracting discriminative representations of forgery cues.
Facial images with fine-grained labels are readily available in existing face forgery detection datasets.
Therefore, we propose UVIF, i.e., Unified Video and Image representation for face Forgery detection.
UVIF processes video and image inputs within a single model and jointly addresses both tasks in a multi-task paradigm, as shown in Figure~\ref{figure:motivation}. Image forgery detection serves as an auxiliary task that provides fine-grained supervision, aiming to improve representation learning for partially forged videos.

The contributions of this paper include:

1) We propose UVIF, a unified framework that utilizes annotated images to provide fine-grained supervision for detecting partially forged videos. It uses a shared backbone and temporal fusion modules to process video and image inputs.

2) A pseudo labeling process is designed to compensate for the absence of frame-level supervision. 
It generates frame-level pseudo labels from image classifier to bridge the representation of video frames and static images.

3) A video-oriented feature alignment strategy is introduced to reduce the distribution gap between images and video frames. It performs feature interpolation using image anchors to adapt the image classifier to video-frame features.

4) Extensive experiments show that UVIF outperforms state-of-the-art methods in partially forged video detection. Ablation studies further demonstrate its data efficiency, with 10k annotated images sufficient to achieve a significant performance boost, as well as its effectiveness across video and image datasets with distinct manipulation types.

This paper extends our previous study presented at \texttt{ECAI 2024} \cite{uvif_liu2024} through the following substantial improvements: 1) We provide a theoretical analysis of existing image-based and video-based methods for partially forged video detection and refine the methodology with clearer notation. 2) We introduce a new video-oriented feature alignment strategy to reduce the distribution gap between video and image datasets containing different manipulation types. 3) We conduct a more comprehensive evaluation of UVIF, including additional ablation studies, backbones, datasets, and analyses.

\section{Related Work}
\subsection{Face Forgery Detection}
Face forgery detection aims to detect forged or synthesized faces in images and videos, which is crucial for the authenticity of visual information that we see every day.
Early studies \cite{mesonet_afchar2018, faceforensics++_rossler2019} formulated face forgery detection as an image binary classification task and employed convolutional neural networks (CNNs) to process facial images.
Subsequent image-based methods further exploited spatial forgery cues, including local textures \cite{madd_zhao2021}, frequency domain \cite{frequency_qian2020}, and inconsistency information \cite{self_blended_shiohara2022, recce_cao2022}.
Recent studies \cite{ hierarchical_liu2024,narrowing_yu2024,effort_yan2025orthogonal,deepfake_xu2026} have also focused on improving robustness and generalization to unseen manipulation methods.
It is worth noting that some image-based methods \cite{frequency_qian2020,  self_blended_shiohara2022, recce_cao2022} have been extended to video face forgery detection. They converted facial videos to individual image frames and assigned a classification label to each frame during training.
However, these methods assumed that all frames in a facial video are the same kind, i.e., all as real, or all as fake, making them unsuitable to process partially forged videos.

As temporal inconsistencies across video frames provide important artifact cues, many video-based methods\cite{ictu_li2018,lips_haliassos2021,biological_ciftci2020,stil_gu2021, forgerynet_he2021, ftcn_zheng2021,  altfreezing_wang2023, tall_xu2023, mintime_coccomini2024, pwtf_kim2025,dip_nie2025, dfd_fcg_han2025} have been proposed for face forgery detection. These methods directly process facial video sequences to model temporal dependencies.
Early studies focused on mining temporal clues by using prior knowledge, such as eye blinks \cite{ictu_li2018}, lip motions \cite{lips_haliassos2021}, and biological signals \cite{biological_ciftci2020}.
Some methods \cite{stil_gu2021, ftcn_zheng2021, altfreezing_wang2023, tall_xu2023, mintime_coccomini2024} directly processed facial video clips using CNN- or transformer-based architectures to learn spatiotemporal representations, achieving better accuracy than image-based methods.
Recent works \cite{dip_nie2025, pwtf_kim2025, dfd_fcg_han2025} have focused on modeling inter-frame relationships to learn discriminative temporal representations.
Despite their swift progress, partially forged video detection still requires further exploration.

\subsection{Multiple Instance Learning}
Multiple instance learning (MIL) \cite{miml_deep_feng2017,mil_attention_ilse2018} is a form of weakly supervised learning with broad applications in medical imaging and video analysis. 
In MIL, a bag of instances is annotated with a single bag-level label, while the exact label for each instance is unavailable.
For a binary classification, negative bags only contain negative instances, while positive bags can contain both positive and negative instances.
Early studies \cite{miml_deep_feng2017} focused on extracting and aggregating features of each instance in a bag via deep neural networks and pooling operations.
Recent approaches \cite{mil_attention_ilse2018, mil_dual_li2021, transmil_shao2021, acmil_zhang2024} have also been developed based on attention mechanisms and transformers.
Li et al. \cite{mil_sharp_li2020} proposed a sharp MIL, i.e., S-MIL, for face forgery detection to handle the problem of partially manipulated faces in videos.
It treated facial frames and videos as instances and bags in MIL, respectively, and designed a sharp loss emphasizing hard instances to address the partially forged videos.
Nevertheless, the performance of MIL methods is limited due to the lack of instance-level labels during training. In this paper, we propose to address the partially forged video detection task from a new perspective, i.e., by incorporating additional image instances with annotations to compensate for the absence of fine-grained supervision information.

\subsection{Unified Architecture Design}
The unified architecture design \cite{clip_radford2021, omnivore_girdhar2022} has gained significant attention recently. It can process input data from different modalities or perform multiple tasks using a single model.
As video and image data are related in structure, i.e., a video can be viewed as a sequence of images, some methods \cite{tsm_lin2019, videoswin_liu2022} have introduced temporal modeling into CNN or transformer backbones originally designed for 2D images and adapted them to video tasks.
Other methods \cite{omni_duan2020, efficient_cao2025, vlab_he2025} investigated using image data and pretrained knowledge to assist video tasks and enhance video representation learning.
Given the strong generalization capability of transformers, some methods \cite{clip_radford2021, imagebind_girdhar2023} used transformers to encode or align data from multiple modalities, such as image, video, audio, and text, into a unified feature space.
Due to diverse training data and the scalability of transformers, these methods can learn effective feature representations across modalities. 
The unified architecture design has demonstrated promising results in other visual tasks, which inspires us to apply it to partially forged video detection by incorporating annotated images.

\begin{figure*}[t]
    \centering
    \includegraphics[width=0.7\linewidth]{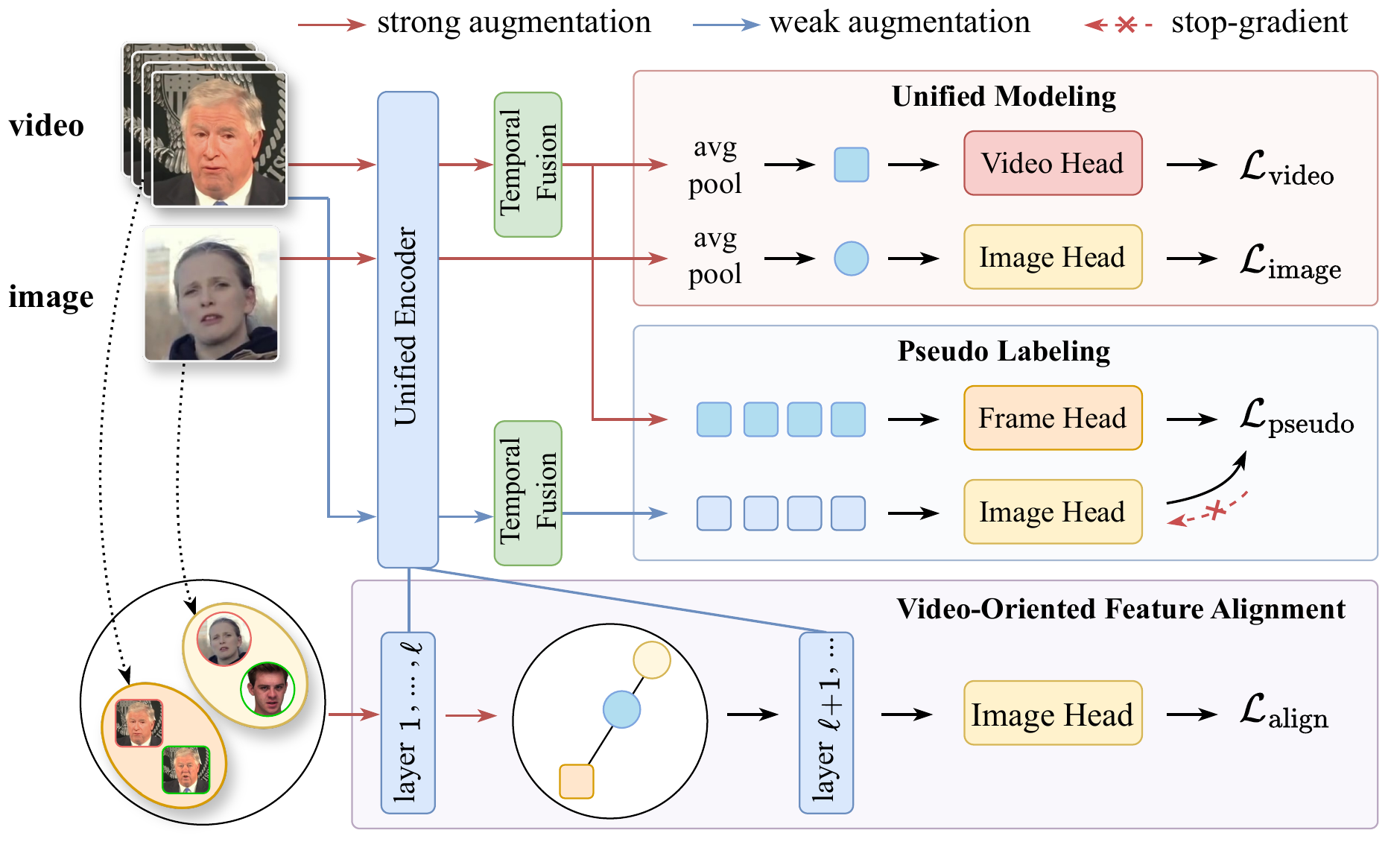}
    \vspace{-1.25em}
    \caption{Overview of the proposed UVIF framework, which comprises three components: 1) unified video and image modeling, where a unified encoder extracts features from both facial videos and images, with temporal fusion applied only to videos, under a multi-task learning paradigm, 2) frame-level pseudo labeling process transfers fine-grained image supervision to video frames, and 3) video-oriented feature alignment interpolates features to reduce the distribution gap between images and video frames. The image head is shared across all three components. During inference, only the unified encoder and video head are used.}
    \vspace{-1em}
    \label{figure:framework}
\end{figure*}

\section{Preliminary}

This section presents the problem formulation and analyzes the limitations of existing video forgery detection methods.

\subsection{Problem Formulation}
\label{section:problem_formulation}
This paper focuses on video face forgery detection, formulated as a binary classification task that determines whether a video clip contains fake faces.

Let $V=\{v_t\}_{t=1}^{T}$ represent a facial video clip, which consists of a sequence of video frames $v_t$, and $T$ is the number of frames. $Y$ represents the binary classification label of the entire video, where $Y \in \{0, 1\}$ denotes whether the video is real or fake, respectively. 
The objective of video face forgery detection is to learn a model that accurately predicts the binary label for a given video. 

For a real video clip, every frame is genuine, whereas the video is labeled as fake if one or more frames are manipulated.
Assuming that there are binary frame labels $\mathbf{y}_{f}=\{y_t\}_{t=1}^{T}$ for each video frame, where $y_t \in \{0, 1\}$, for $t = 1, 2, ..., T$, this premise can be described as:
\begin{equation}
\label{eq:problem_formula}
Y =
\begin{cases}
1, & \exists\, y_t=1,\\
0, & \text{otherwise}.
\end{cases}
\end{equation}
This formulation corresponds to the setting of multiple instance learning (MIL) \cite{miml_deep_feng2017,mil_attention_ilse2018}. Specifically, the video frames $V=\{v_t\}_{t=1}^{T}$ can be treated as a bag of instances in MIL, and $Y$ serves as the bag-level label.
During training, only the label $Y$ is used, while the individual labels $\mathbf{y}_{f}=\{y_t\}_{t=1}^{T}$ of each frame are not available.

\subsection{Limitation of Existing Methods}

Most previous studies assume that all frames in a fake video are manipulated. However, this assumption does not hold for partially forged videos, and applying these methods can lead to unreliable results.

\subsubsection{Image-based Methods}

Given a video $V=\{v_t\}_{t=1}^{T}$ with a video-level label $Y$, image-based methods treat each frame as an individual image for binary classification and construct a frame-level training set by assigning $Y$ to every frame:
\begin{equation}
\widetilde{\mathcal{D}}_f
=
\{(v_t,Y)\}_{t=1}^{T}.
\end{equation}

The correct frame-level training set is instead given by:
\begin{equation}
\mathcal{D}_f
=
\{(v_t,y_t)\}_{t=1}^{T}.
\end{equation}

When a video is real or fully forged, each constructed label is consistent with the true frame-level label, i.e., $Y=y_t$. 
However, for a partially forged video, Eq.~\ref{eq:problem_formula} implies that some real frames satisfy $y_t=0$ while $Y=1$. Therefore, the constructed frame training set differs from the correct one:
\begin{equation}
\widetilde{\mathcal{D}}_f \neq \mathcal{D}_f,
\end{equation}
indicating that real frames are incorrectly treated as fake samples during training. Consequently, these incorrect labels introduce noise into frame-level classification.

\subsubsection{Video-based Methods}

Let $z_t$ denote the logit for frame $v_t$. Video-based methods aggregate the frame-level logits using average pooling to obtain the video-level logit $z$:
\begin{equation}
z=\frac{1}{T}\sum_{t=1}^{T} z_t,\quad p=\mathrm{sigmoid}(z),
\end{equation}
where $p$ denotes the video-level probability. The binary cross-entropy loss is defined as:
\begin{equation}
\mathcal{L} = \mathcal{H}(y,p)
=
-y\log p-(1-y)\log(1-p).
\end{equation}

For a fake video, where $y=1$, the gradient with respect to each frame-level logit is:
\begin{equation}
\frac{\partial \mathcal{L}}{\partial z_t}
=
\frac{\partial \mathcal{L}}{\partial p}
\frac{\partial p}{\partial z}
\frac{\partial z}{\partial z_t}
=
-\frac{1-p}{T}<0.
\end{equation}

Thus, all frames in a partially forged video are optimized toward the fake class, pushing real frames toward the fake distribution as well. This distorts frame-level representation learning, especially when forged frames are sparse.

\subsection{Our Approach}

To address these limitations, we introduce an additional set of annotated facial images to provide fine-grained supervision. Let $(s,y_s)\sim\mathcal{D}_s$ denote a facial image and its binary label, where $y_s\in\{0,1\}$. Compared with the video-level label $Y$, the image label $y_s$ is more fine-grained and precise, and annotated facial images are readily available in face forgery datasets. As a static facial image can be regarded as an individual video frame, samples from $\mathcal{D}_s$ can substitute for the unavailable frame-level annotations $(v_t,y_t)\sim\mathcal{D}_f$. This motivates us to learn unified representations for videos and images.

\section{Methodology}
An overview of the proposed UVIF is presented in Figure~\ref{figure:framework}.
This section is organized as follows. Section~\ref{section:unifed_modeling} introduces unified video and image modeling as the foundation of the framework. Section~\ref{section:pseudo_labeling} describes the pseudo labeling strategy that transfers image supervision to video frames. Section~\ref{section:feature_alignment} presents the video-oriented feature alignment. Section~\ref{section:optimization} details the overall optimization objective.

\subsection{Unified Video and Image Modeling}
\label{section:unifed_modeling}
Partial face forgery detection is limited by the absence of frame-level annotations. We introduce an annotated facial image set to provide fine-grained supervision. The model therefore needs to process both videos and images during training.
Inspired by the popular unified architecture designs \cite{tsm_lin2019,videoswin_liu2022,omnivore_girdhar2022}, we use a unified encoder $\phi(\cdot)$ with shared parameters to extract features for both video and image inputs.
The video and image inputs are represented as a 4D tensor $X \in \mathbb{R}^{T\times 3 \times H \times W}$, where $H$ and $W$ denote spatial dimensions, and $T$ denotes temporal dimension, with $T\!=\!1$ for images.
The unified encoder uses typical 2D backbones to extract spatial features from images and video frames in the same manner.
Temporal fusion is applied to video inputs to compensate for the lack of temporal interaction across video frame features. 
After average pooling, the unified encoder produces feature vectors $z_v,z_s\in\mathbb{R}^{C}$, where $C$ denotes the number of channels.

Two separate linear classification heads are applied to make predictions for videos and images.
During training, each mini-batch is constructed by randomly sampling a set of facial videos and images.
The classification losses are defined as:
\begin{equation}
\mathcal{L}_\mathrm{video}=\mathcal{H}(Y,p_v),
\quad
\mathcal{L}_\mathrm{image}=\mathcal{H}(y_s,p_s),  
\end{equation}
where $p_v$ and $p_s$ denote the predicted probabilities for video and image inputs, respectively, and $\mathcal{H}(\cdot)$ denotes the cross-entropy loss.
The unified encoder is thus optimized by both video and image supervision.
Apart from video face forgery detection, it also learns to extract discriminative features for determining the authenticity of facial images.

The unified encoder is implemented with typical 2D CNN \cite{resnet_he2016, convnext_woo2023} or transformer \cite{videoswin_liu2022} backbones. For temporal fusion, we adopt the Temporal Shift Module (TSM) \cite{tsm_lin2019} for CNN backbones and temporal attention \cite{videoswin_liu2022} for transformers.

\begin{figure}[!t]
    \centering
    \vspace{-1.5em}
    \includegraphics[width=0.65\linewidth]{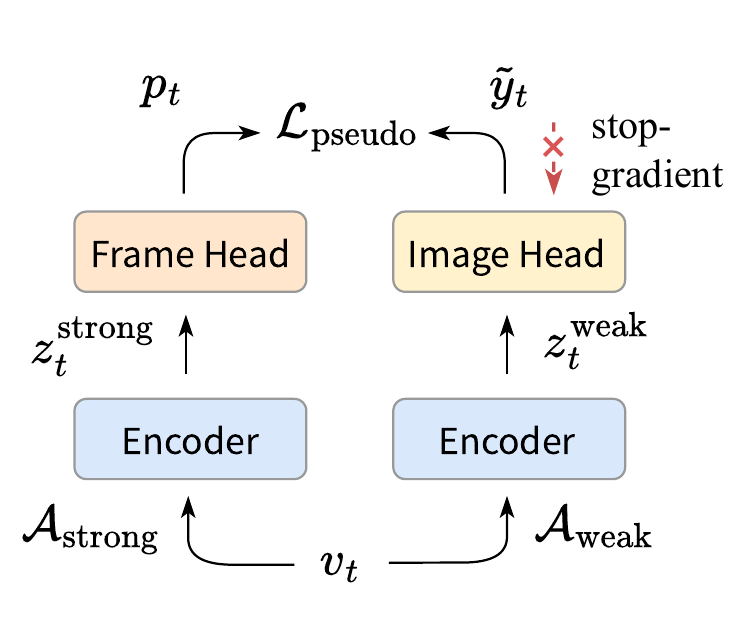}
    \vspace{-1.5em}
    \caption{Illustration of the pseudo labeling. For each video frame $v_t$, the image classification head predicts the pseudo label $\tilde y_t$ from a weakly augmented view, while the frame head predicts $\hat p_t$ from a strongly augmented view.}
    \label{figure:pseudo_labeling}
\end{figure}

\subsection{Bridging Video and Image Representation}
\label{section:pseudo_labeling}
The unified modeling process uses annotated samples from the facial image set $(s,y_s)\sim\mathcal{D}_s$ as substitutes for the original video frame set $(v_t,y_t)\sim\mathcal{D}_f$.
As the frame-level labels $y_t$ of video samples are unavailable during training, the model cannot receive direct supervision from each video frame.
Since $\mathcal{D}_s$ may differ from $\mathcal{D}_f$ due to limited number of image samples or distribution differences between images and video frames, the model may learn biased representations from images that mismatch with video frames.
To address this issue, we introduce an auxiliary pseudo labeling process for facial videos. 
It generates pseudo labels for video frames using the image classification head, aiming to bridge the representation of video frames and images.

As shown in Figure~\ref{figure:pseudo_labeling}, we generate two augmented views for each video frame $v_t$:
\begin{equation}
v_t^\mathrm{weak}=\mathcal{A}_\mathrm{weak}(v_t),\quad
v_t^\mathrm{strong}=\mathcal{A}_\mathrm{strong}(v_t),
\end{equation}
where $\mathcal{A}_\mathrm{weak}$ denotes weak augmentation, including resizing, cropping, and horizontal flipping, while $\mathcal{A}_\mathrm{strong}$ denotes strong augmentation with additional image compression and color perturbation.
The two views are fed into the unified encoder to obtain the corresponding feature vectors $z_t^\mathrm{weak}$ and $z_t^\mathrm{strong}$.
The image classification head predicts the soft pseudo label $\tilde{y}_t$ from the weakly augmented view, while a newly introduced frame head predicts the frame-level probability $p_t$ from the strongly augmented view. The frame head has the same architecture as the image head but uses unshared parameters.

The pseudo labeling loss is defined as:
\begin{equation}
\mathcal{L}_\mathrm{pseudo}
=
\frac{1}{T}
\sum_{t=1}^{T}
\mathcal{H}
\left(
\operatorname{stopgrad}(\tilde{y}_t),
p_t
\right),
\end{equation}
where the stop-gradient operation $\mathrm{stopgrad(\cdot)}$ is applied to pseudo labels $\tilde{y}_{t}$ to avoid the collapse of training.

The pseudo labeling process follows semi-supervised learning paradigms \cite{ fixmatch_sohn2020, debiased_chen2022}, which generate reliable pseudo labels from a weakly augmented view and provide supervision to a strong view. Since the frame head is discarded at inference, it only affects representation learning in the encoder.

\subsection{Video-Oriented Feature Alignment}
\label{section:feature_alignment}

The unified modeling and pseudo labeling processes introduce image-level and frame-level supervision to learn discriminative representations for distinguishing authentic and manipulated video frames. 
However, both processes rely on the image classification head to provide effective supervision.
Static images and video frames may contain different manipulation types, which may cause an image head trained on images to produce unreliable predictions for video frames.

\begin{figure}[!t]
    \centering
    \vspace{-0.5em}
    \includegraphics[width=0.9\linewidth]{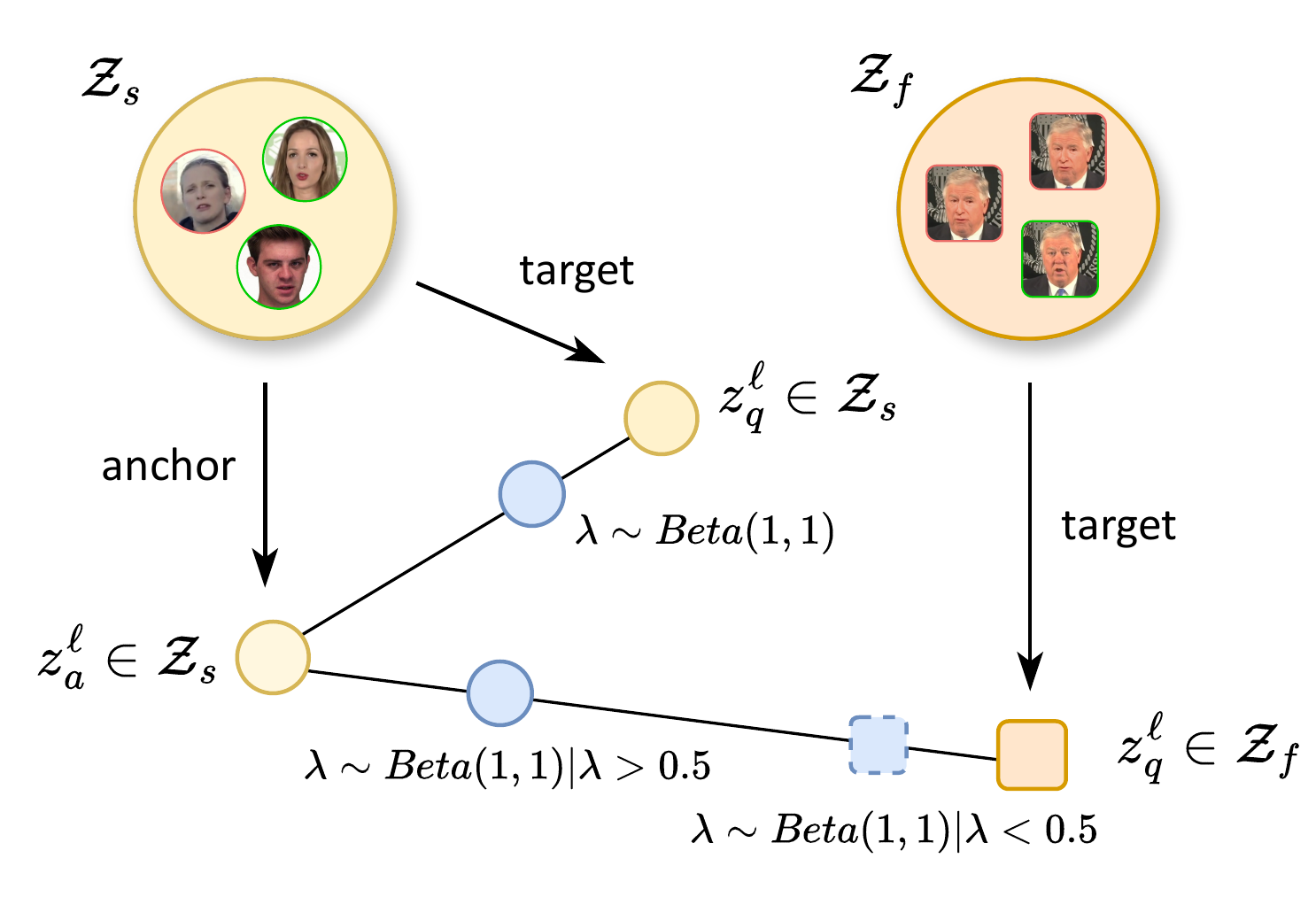}
    \vspace{-1.5em}
    \caption{Illustration of video-oriented feature alignment. The image feature set $\mathcal{Z}_{s}$ and video frame set $\mathcal{Z}_{f}$ are first constructed. An image feature serves as the anchor $z_a^{l}$ and is interpolated with a target feature $z_q^{l}$. For image-video interpolation, $\lambda>0.5$ makes the interpolated feature closer to the image.}
    \label{figure:feature_alignment}
\end{figure}

To address this issue, we propose a feature alignment strategy that adapts the image head to video frame representations. Specifically, we perform feature-level interpolation between image and video frame features. The image feature serves as the anchor, and the interpolation is biased toward it to keep the mixed feature compatible with the image head, as shown in Figure~\ref{figure:feature_alignment}. This strategy regularizes the image head in the intermediate feature space and encourages a smoother transition between the image and video frame distributions.

Given an encoder layer $\ell$ randomly sampled from the candidate index set $\mathcal{I}$, we construct two feature sets for images and video frames, $\mathcal{Z}_{s}$ and $\mathcal{Z}_{f}$, by extracting intermediate features through a forward pass from input feature to layer $\ell$, denoted by $\phi_{0:\ell}(\cdot)$. For typical backbones\cite{videoswin_liu2022, resnet_he2016,convnext_woo2023}, $\ell$ ranges from 0 to 4, with $\ell\!=\!0$ denoting the input feature.

The image feature set $\mathcal{Z}_{s}$ is defined as:
\begin{equation}
\label{eq:image_set}
\mathcal{Z}_{s}
=
\left\{
z_{s,i}^{\ell}
\mid s_i\in\mathcal{D}_s
\right\}.
\end{equation}

The video frame feature set $\mathcal{Z}_{f}$ includes both real and fake frames. Every frame from real videos is treated as a real sample, while fake frames from fake videos are selected when their pseudo-label confidence exceeds the threshold $\tau$:
\begin{equation}
\label{eq:video_set}
\mathcal{Z}_{f}
=
\left\{
z_{t}^{\ell}
\mid
v_t \in V,\ 
Y=0\vee(Y=1 \wedge \tilde{y}_{t,1}>\tau)
\right\}.
\end{equation}

We construct interpolated samples by using an image feature $z_a^{\ell}$ as the anchor and randomly selecting a target feature $z_q^{\ell}$ from the image or video frame set:
\begin{equation}
z_a^{\ell}\in\mathcal{Z}_{s},
\quad
z_q^{\ell}\in\mathcal{Z}_{s}\cup\mathcal{Z}_{f}.
\end{equation}
Note that the anchor feature $z_a^{\ell}$ is restricted to $\mathcal{Z}_{s}$ rather than $\mathcal{Z}_{s}\cup\mathcal{Z}_{f}$ to avoid generating features dominated by video frames, which may introduce noise to the image head.

The interpolated feature and its soft label are defined as:
\begin{equation}
z_{m}^{\ell}
=
\lambda z_a^{\ell}+(1-\lambda)z_q^{\ell},
\quad
y_{m}
=
\lambda y_a+(1-\lambda) y_q,
\end{equation}
where $y_a$ and $y_q$ denote the corresponding labels. Specifically, $y_q$ is a pseudo label when $z_q^{\ell}$ is a video frame feature.

The weight $\lambda$ is sampled according to the source of the target feature:
\begin{equation}
\label{eq:lambda}
\lambda \sim
\begin{cases}
\mathrm{Beta}(1,1), 
& z_q^{\ell}\in\mathcal{Z}_{s},\\
\mathrm{Beta}(1,1)\mid \lambda > 0.5,
& z_q^{\ell}\in\mathcal{Z}_{f}.
\end{cases}
\end{equation}
For frame target features $z_q^{\ell}\in\mathcal{Z}_{f}$, we truncate the distribution by enforcing $\lambda > 0.5$, 
biasing the interpolated feature toward the image anchor for compatibility with the image head.

We construct $N_\mathrm{align}$ interpolated samples at each iteration.
The interpolated features are passed through the remaining encoder layers and then predicted by the image head. The alignment loss is defined as:
\begin{equation}
\mathcal{L}_{\mathrm{align}}=\gamma\mathcal{H}(y_{m},h_s(\phi_{\ell+1:}(z_{m}^{\ell}))),
\end{equation}
where $\phi_{\ell+1:}(\cdot)$ denotes the forward pass from layer $\ell+1$ to the output feature, $h_s(\cdot)$ denotes the image head, and $\gamma$ denotes the loss weight. 
For hyperparameters, we use $\mathcal{I}\!=\!\{0,1,2\}$, $\tau\!=\!0.9$, $N_\mathrm{align}\!=\!64$, and $\gamma\!=\!0.5$ in the experiments.

This strategy is related to manifold mixup \cite{manifold_verma2019}, as both perform interpolation in feature space. However, our strategy interpolates features from different distributions and biases the interpolated features toward image anchors, enabling the image head to adapt to video frame features while preserving its classification ability on both images and video frames.

\subsection{Optimization}
\label{section:optimization}

The overall framework is optimized end-to-end under a multi-task learning paradigm using the following objective:
\begin{equation}
    \mathcal{L} = \mathcal{L}_{\mathrm{video}} +   \mathcal{L}_{\mathrm{image}} +  \mathcal{L}_{\mathrm{pseudo}} + \mathcal{L}_{\mathrm{align}},
\end{equation}
where $\mathcal{L}_{\mathrm{video}}$, $\mathcal{L}_{\mathrm{image}}$, $\mathcal{L}_{\mathrm{pseudo}}$, and $\mathcal{L}_{\mathrm{align}}$ denote the video classification, image classification, pseudo labeling, and feature alignment losses, respectively. 
Note that the proposed method incurs no additional computational overhead during inference as only the video classification process is applied.

\section{Experiments}

\subsection{Experimental Setup}
\subsubsection{Datasets and Evaluation Metrics}
We conduct experiments on publicly available face forgery detection datasets ForgeryNet \cite{forgerynet_he2021} and DFDC (preview) \cite{dfdc_dolhansky2020}.

ForgeryNet is a large-scale face forgery detection dataset with over 220k facial video clips and 2.9m static images from over 5k subjects. It contains 15 facial manipulation approaches and over 36 mix-perturbations. Most fake videos are partially manipulated, making this dataset challenging for face forgery detection. For video forgery classification, we follow \cite{forgerynet_he2021} and use about 140k videos for training and 18k for validation. For the image set, ForgeryNet includes over 2.3m training images, but unless otherwise specified, we use only a randomly selected subset of 100k images (less than 5\%).

DFDC is the preview dataset for the Deepfake Detection Challenge \cite{dfdc_dolhansky2020}, which includes 1131 real videos and 4113 forged videos generated using two synthesis methods. Most videos are partially forged. We follow the original dataset split in \cite{dfdc_dolhansky2020} and use 4464 videos for training and 780 for testing.

For evaluation metrics, we follow \cite{forgerynet_he2021, dfdc_dolhansky2020} and adopt video-level Accuracy (Acc) and Area under the ROC curve (AUC) to evaluate the performance of video forgery detection.

\subsubsection{Implementation Details}
We adopt the Swin Transformer \cite{videoswin_liu2022} or CNNs \cite{resnet_he2016,convnext_woo2023} with TSM \cite{tsm_lin2019} as representative implementations of our unified encoder. Following \cite{forgerynet_he2021}, we use RetinaFace \cite{retinaface_deng2020} to extract facial regions and resize them to 224$\times$224. We sample 32 frames with a temporal stride of 4 from each video clip for training and use the center clip for evaluation. We employ both weak and strong data augmentation: weak augmentation includes random resizing, cropping, and flipping, while strong augmentation further includes compression and color perturbation. We apply strong augmentation to all compared methods, whereas UVIF uses both augmentations to construct two views for pseudo labeling. The video batch size is 32, and UVIF also randomly samples 256 annotated images per iteration without requiring the same subjects across videos and images. All models are trained on four NVIDIA Tesla V100 GPUs with a learning rate of 0.01, momentum of 0.9, and weight decay of 0.0001. We train the models for 50K iterations on ForgeryNet and 20K iterations on DFDC to ensure convergence. Unless otherwise specified, all compared methods use the same hyperparameters.

\begin{table}[!t]
    \centering
    \caption{
    Comparison with the state-of-the-art methods on ForgeryNet \cite{forgerynet_he2021}.
    Results marked with $\dagger$ are cited from the original papers.
    All the remaining results are reimplemented using the same protocol for a fair comparison.
    The \#params denotes the number of parameters. The FLOPs are measured under spatial size 224 $\times$ 224 and 32 frames. Bold indicates the best results.
    }
    \label{table:sota}
    \setlength{\tabcolsep}{3pt}
    \begin{tabular}{c|llcccc}
    \toprule
    \multicolumn{2}{l}{Method} & Backbone 
    & \makecell{\footnotesize \#params \\ \footnotesize (M)} 
    & \makecell{\footnotesize FLOPs \\ \footnotesize (G)} 
    & Acc & AUC \\ 
    \midrule
        \multirow{6}{*}{\rotatebox[origin=c]{90}{\makecell[c]{Image-based}}} & Xception \cite{xception_chollet2017} & Xception & 21 & 146 & 66.66 & 72.06 \\
        & EfficientNet \cite{efficientnet_tan2019} & Efficient-B4 & 18 & 49 & 70.50 & 77.04 \\
        & Swin \cite{videoswin_liu2022} & Swin-T & 28 & 144 & 71.72 & 77.00 \\
        & Swin \cite{videoswin_liu2022} & Swin-S & 49 & 281 & 71.46 & 77.80 \\
        & CLIP-ViT \cite{clip_radford2021} & ViT-B & 86 & 18 & 70.63 & 77.11 \\
        & Effort \cite{effort_yan2025orthogonal} & ViT-L & 303 & 2598 & 72.51 & 80.64 \\
        \midrule
        \multirow{16}{*}{\rotatebox[origin=c]{90}{Video-based}}
        & TSM \cite{forgerynet_he2021} $^\dagger$ & R-50 & 24 & 132 & 88.04 & 93.05 \\
        & SlowFast \cite{forgerynet_he2021} $^\dagger$ & R3D-50  & 34 & 51 & 88.78 & 93.88 \\ 
        & TSM \cite{tsm_lin2019} & R-50 & 24 & 132 & 83.60 & 90.12 \\
        & SlowFast \cite{slowfast_feichtenhofer2019} & R3D-50 & 34 & 51 & 84.04 & 91.02 \\
        & SlowFast \cite{slowfast_feichtenhofer2019} & R3D-101 & 62 & 97 & 84.31 & 90.87 \\
        & VideoSwin \cite{videoswin_liu2022} & Swin-T & 28 & 88 & 83.89 & 90.53 \\
        & VideoSwin \cite{videoswin_liu2022} & Swin-S & 50 & 166 & 84.56 & 91.19 \\
        & TimeSformer \cite{timesformer_bertasius2021} & ViT-B & 86 & 563 & 84.17 & 90.18 \\
        & UniFormer \cite{uniformer_li2022} & UniFormer-S & 21 & 110 & 84.01 & 89.33 \\
        & FTCN \cite{ftcn_zheng2021} & R3D-50 & 57 & 68 & 78.62 & 85.00 \\
        & STIL \cite{stil_gu2021} & SCNet-50 & 23 & 151 & 83.55 & 88.98 \\
        & AltFreezing \cite{altfreezing_wang2023} & R3D-50 & 27 & 114 & 82.80 & 85.73 \\
        & TALL \cite{tall_xu2023} & Swin-B & 87 & 16 & 82.17 & 88.12 \\
        & PwTF \cite{pwtf_kim2025} & Designed & 197 & 90 & 75.98 & 81.92 \\
        & MINTIME-XC \cite{mintime_coccomini2024} $^{\dagger}$  & Xception & 103 & 223 & 87.64 & 94.25 \\
        & DFD-FCG \cite{dfd_fcg_han2025} & ViT-L & 306 & 2624 & 84.13 & 91.73 \\
        \midrule
        \multirow{5}{*}{\rotatebox[origin=c]{90}{MIL}}
        & MIL \cite{mil_attention_ilse2018} & Swin-T & 28 & 88 & 84.10 & 91.31 \\
        & S-MIL \cite{mil_sharp_li2020} & Swin-T & 28 & 88 & 83.95 & 91.20 \\
        & TransMIL \cite{transmil_shao2021} & Swin-T & 30 & 96 & 82.89 & 87.38 \\
        & DSMIL \cite{mil_dual_li2021} & Swin-T & 28 & 88 & 83.84 & 90.59 \\
        & ACMIL \cite{acmil_zhang2024} & Swin-T & 28 & 88 & 83.74 & 90.96 \\
        \midrule
        \multirow{3}{*}{\rotatebox[origin=c]{90}{Ours}}
        & UVIF (\textbf{Ours}) & R-50 & 24 & 132 & 85.51 & 93.12 \\
        & UVIF (\textbf{Ours}) & Swin-T & 28 & 88 & 88.02 & 95.10 \\ 
        & UVIF (\textbf{Ours}) & Swin-S & 50 & 166 & \textbf{88.94} & \textbf{95.69} \\
        \bottomrule
    \end{tabular}
\end{table}

\subsection{Comparison to State-of-the-art}

We evaluate the performance of our proposed methods on the ForgeryNet \cite{forgerynet_he2021} dataset, and compare them with state-of-the-art methods, including image-based and video-based methods for video forgery detection, as well as typical multiple instance learning methods, as detailed in Table \ref{table:sota}.
The image-based methods are initialized with pre-trained weights on ImageNet-1k \cite{imagenet_russakovsky2015}, while video-based methods are initialized with pre-trained weights on Kinetics-400 \cite{kinetics_kay2017} for better performance.
The TSM \cite{tsm_lin2019} and VideoSwin \cite{videoswin_liu2022} methods can be viewed as video-based baselines for UVIF, as they employ equivalent architectures during evaluation.

The comparison results reveal a clear performance gap between image-based and video-based methods. The best image-based method Effort~\cite{effort_yan2025orthogonal} achieves 72.51\% accuracy and 80.64\% AUC, whereas most video-based methods perform substantially better. This gap indicates that image-based methods cannot model temporal cues and are more susceptible to label noise in partially forged videos.
The proposed UVIF methods achieve superior performance over previous methods. The best model, UVIF-Swin-S, achieves 88.94\% accuracy and 95.69\% AUC. Compared with the TSM and VideoSwin baselines, the corresponding UVIF variants introduce no additional parameters or FLOPs, and UVIF-Swin-T achieves a trade-off between detection accuracy and computational efficiency.
Besides, MIL methods \cite{mil_attention_ilse2018, mil_sharp_li2020, mil_dual_li2021} are also implemented using the VideoSwin-T architecture, and each input video clip is treated as a bag in MIL. The results in Table \ref{table:sota} indicate that although MIL methods can improve the AUC to some extent by grouping similar video frames, the performance gains remain limited. In contrast, our UVIF methods can significantly improve detection performance using fine-grained image annotations.

We also compare UVIF with existing methods on the DFDC \cite{dfdc_dolhansky2020} dataset, as presented in Table \ref{table:sota_dfdc}. UVIF is trained using the selected image set from ForgeryNet.
It significantly enhances detection accuracy over the TSM \cite{tsm_lin2019} and VideoSwin \cite{videoswin_liu2022} baselines, e.g., +3.99\% accuracy for Swin-T. Our method also outperforms state-of-the-art methods on DFDC.

\begin{table}[!t]
    \centering
    \caption{Results on the DFDC \cite{dfdc_dolhansky2020}. Results marked with $\dagger$ are cited from original papers. Bold indicates the best results.}
    \label{table:sota_dfdc}
    \begin{tabular}{lcc}
        \toprule
        Method & Acc & AUC \\ 
        \midrule
        Xception-avg \cite{faceforensics++_rossler2019} $^\dagger$ & 84.58 & -  \\
        S-MIL \cite{mil_sharp_li2020} $^\dagger$ & 83.78 & - \\
        S-MIL-T \cite{mil_sharp_li2020} $^\dagger$ & 85.11 & - \\
        STIL \cite{stil_gu2021} $^\dagger$ & 89.80 & - \\
        TSM-R50 \cite{tsm_lin2019} & 81.08 & 90.67 \\
        SlowFast-R50 \cite{slowfast_feichtenhofer2019} & 83.53 & 92.10 \\
        VideoSwin-T \cite{videoswin_liu2022} & 84.68 & 92.72 \\
        VideoSwin-S \cite{videoswin_liu2022} & 88.29 & 94.85 \\
        Uniformer-S \cite{uniformer_li2022} & 85.07 & 94.02 \\
        TimeSformer \cite{timesformer_bertasius2021} & 87.00 & 94.58 \\
        AltFreezing \cite{altfreezing_wang2023} & 82.50 & 89.83 \\
        TALL \cite{tall_xu2023} & 86.23 & 92.89 \\
        PwTF \cite{pwtf_kim2025} & 77.22 & 82.54 \\
        MINTIME-XC \cite{mintime_coccomini2024} $^\dagger$ & 86.40 & 95.20 \\
        DFD-FCG \cite{dfd_fcg_han2025} & 85.20 & 93.39 \\
        \midrule
        UVIF-R50 (\textbf{Ours}) & 86.49 & 92.88 \\
        UVIF-Swin-T (\textbf{Ours}) & 88.67 & 95.26 \\
        UVIF-Swin-S (\textbf{Ours}) & \textbf{90.73} & \textbf{96.05} \\
        \bottomrule
    \end{tabular}
\end{table}

\subsection{Ablation Studies and Analysis}
\subsubsection{Effectiveness of the Core Components}
We first perform ablation experiments to analyze the components of the UVIF framework, as shown in Table \ref{table:components}.
Specifically, UVIF contains four primary components: image supervision $\mathcal{L}_\mathrm{image}$ in unified modeling, temporal fusion for video inputs, $\mathcal{L}_\mathrm{pseudo}$ in pseudo labeling process, and $\mathcal{L}_\mathrm{align}$ in feature alignment.
The first row corresponds to a vanilla Swin-T model trained on videos from ForgeryNet \cite{forgerynet_he2021}.
Applying temporal fusion in row 3 brings significant performance gains (+8.85\% accuracy and +7.67\% AUC), indicating the importance of temporal information in face forgery detection.
Combining images and videos for training (row 4) further improves the accuracy to 86.77\%, showing the benefit of fine-grained image annotations.
Results in row 2 also show that image supervision improves performance even without temporal fusion. This further reflects the effectiveness of annotated images, as the features of images and individual video frames are similar.
Adding the pseudo labeling loss in row 5 further improves the results, indicating its efficacy in bridging the representations of images and video frames.
Finally, incorporating feature alignment in row 6 yields the best performance, with 88.02\% accuracy and 95.10\% AUC, demonstrating its effectiveness in reducing the distribution gap between images and video frames.

\begin{table}[!t]
    \caption{Effectiveness of the core components of UVIF with Swin-T. $\mathcal{L}_{\mathrm{image}}$: image supervision in unified modeling; temporal: temporal fusion for video inputs; $\mathcal{L}_{\mathrm{pseudo}}$: pseudo labeling; $\mathcal{L}_{\mathrm{align}}$: video-oriented feature alignment.}
    \label{table:components}
    \centering
    \begin{tabular}{cccccc}
        \toprule
        $\mathcal{L}_\mathrm{image}$ & Temporal   & $\mathcal{L}_\mathrm{pseudo}$ & $\mathcal{L}_\mathrm{align}$ & Acc & AUC \\
        \midrule
        - & - & - & - & 75.04 & 82.86 \\
        \checkmark & - & - & - & 80.02 & 84.99 \\
        - & \checkmark & - & - & 83.89 & 90.53 \\
        \checkmark & \checkmark & - & - & 86.77 & 94.39 \\
        \checkmark & \checkmark & \checkmark & - & 87.30& 94.59 \\
        \checkmark & \checkmark & \checkmark & \checkmark & \textbf{88.02} & \textbf{95.10} \\
    \bottomrule
    \end{tabular}
\end{table}

\subsubsection{Number of Training Images}
\begin{figure}[!t]
    \centering
    \includegraphics[width=0.9\linewidth]{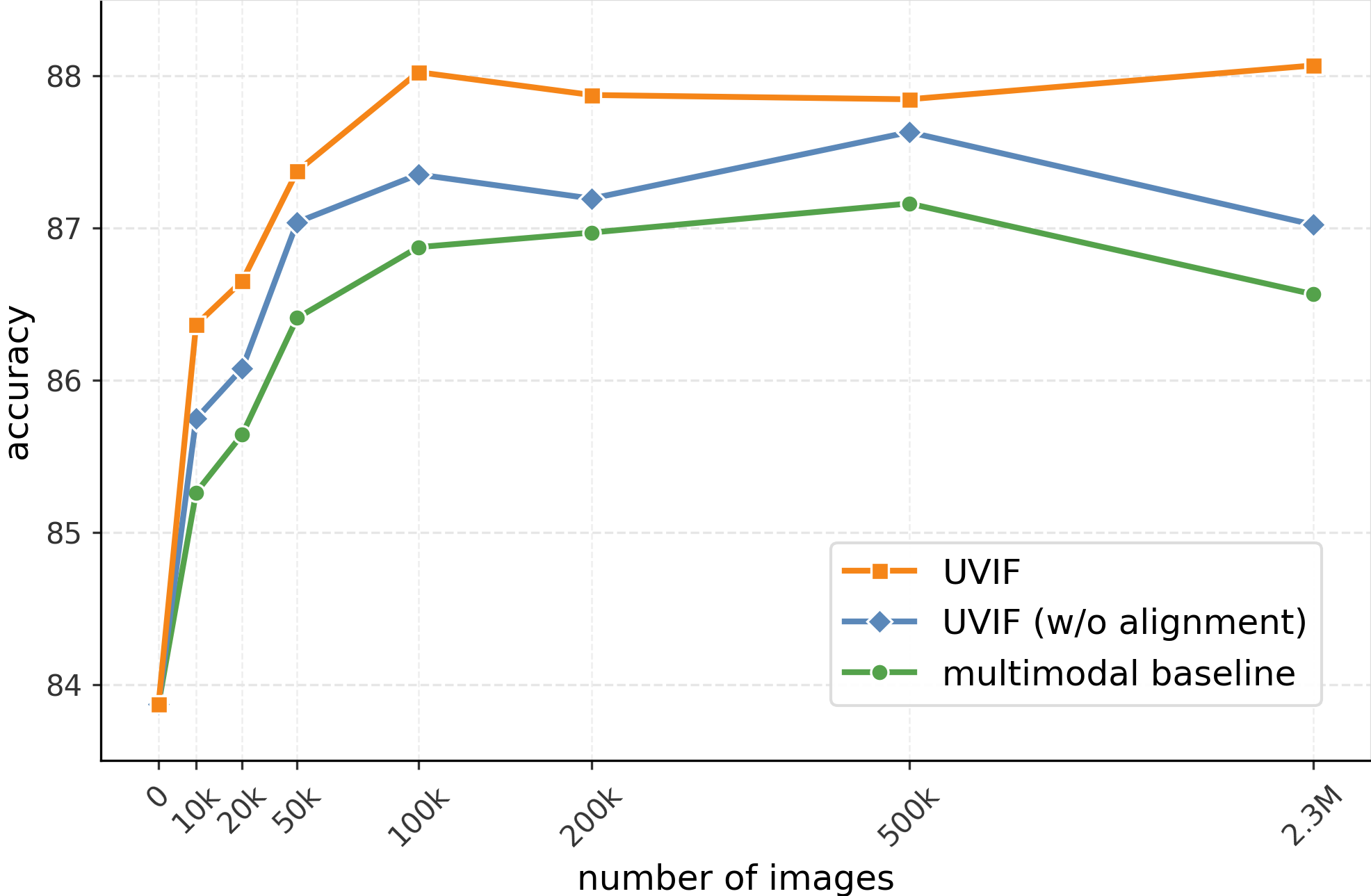}
    \vspace{-0.5em}
    \caption{Ablation on the number of training images under three settings, i.e., multimodal baseline with Swin-T, UVIF without feature alignment, and UVIF.}
    \label{figure:ablation_images}
\end{figure}

We then evaluate the performance of UVIF using different numbers of training images, as illustrated in Figure \ref{figure:ablation_images}.
We randomly sample 10k, 20k, 50k, 100k, 200k, 500k, and 2.3M (all) training images from ForgeryNet \cite{forgerynet_he2021}, and perform experiments under three settings, i.e., UVIF, UVIF without feature alignment, and a multimodal baseline trained jointly on videos and images.
The results show that the accuracy of all methods significantly improves when the training images increase from 0 to 100k and reaches saturation at 100k. Thus, 100k is sufficient for the current video set, and adding more images, i.e., even up to 2.3M, does not provide further gains. 100k images with annotations are easily available from open-access datasets, making UVIF practical for improving partially forged video detection.
Even with a small portion of images, e.g., 10k, UVIF achieves clear performance gains over the video classification baseline (86.36\% vs. 83.89\% accuracy).
This indicates that the UVIF model gradually learns useful supervision information from annotated images, rather than relying on a large number of images to improve its representation.
Meanwhile, the comparison of three curves shows that 1) incorporating pseudo labeling process and feature alignment are always effective using different numbers of training images, and 2) the performance improvement is larger when 100k or more images are used.
This indicates that the model requires a certain number of images to train a reliable image head for pseudo label generation and to learn effective alignment between the distributions of images and video frames.

\subsubsection{Performance on Videos with Different Forged Ratios}
\begin{figure}[!t]
    \centering
    \includegraphics[width=0.9\linewidth]{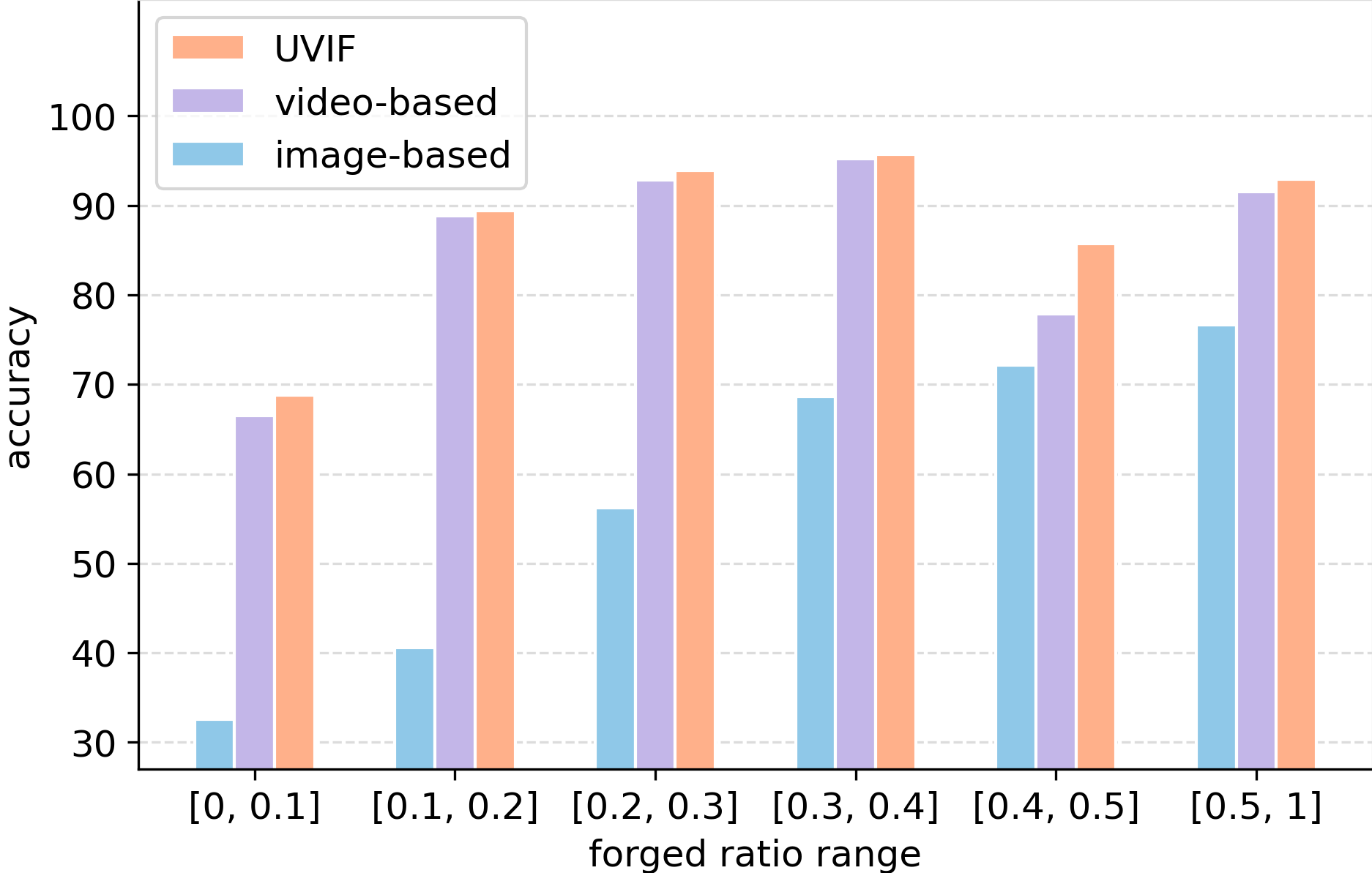}
    \vspace{-0.5em}
    \caption{Detection accuracy on ForgeryNet videos with different forged-frame ratios. The image- and video-based baselines use Swin-T as the backbone.}
    \label{figure:ablation_ratios}
\end{figure}

Figure \ref{figure:ablation_ratios}  illustrates the accuracy achieved on videos with different forged ratios from the ForgeryNet \cite{forgerynet_he2021} validation set.
We utilize the annotations for temporal forgery localization \cite{forgerynet_he2021} task to compute the ratio of forged frames of each partially forged video, and then divide them into six groups based on their forged ratios: [0, 0.1], [0.1, 0.2], [0.2, 0.3], [0.3, 0.4], [0.4, 0.5], and [0.5, 1].
The sample distribution of each group is 0.07, 0.19, 0.22, 0.22, 0.15, and 0.15, respectively.
We compare UVIF with image-based and video-based baselines using Swin-T. 
The results show that the video-based baseline consistently outperforms the image-based baseline, with larger gains when forged frames are sparse.
UVIF consistently enhances the video-based baselines across all forged-ratio groups.
This indicates that UVIF leverages annotated images to learn discriminative representations that distinguish real and forged video frames in partially forged videos.

\begin{table}[t]
\centering
\caption{Ablation study of video-oriented feature alignment. ``Feature Source'' indicates the interpolation feature sets, with $I$ and $V$ representing image and video frame features. ``Image Anchor'' specifies whether the anchor is restricted to images, and ``Bias'' indicates the interpolation direction controlled by $\lambda$.}
\label{table:ablation_alignment}
\resizebox{\linewidth}{!}{
\begin{tabular}{lccccc}
\toprule
Setting & \makecell[c]{Feature\\Source} & \makecell[c]{Image\\Anchor} & Bias & Acc & AUC \\
\midrule
\multicolumn{6}{l}{\textit{\textbf{Baselines}}} \\
video baseline & - & - & - & 83.89 & 90.53 \\
UVIF (w/o alignment) & - & - & - & 87.30 & 94.59 \\
\midrule
\multicolumn{6}{l}{\textit{\textbf{Single-domain interpolation}}} \\
image mixup & I & - & - & 87.44 & 94.54 \\
video mixup & V & - & - & 87.02 & 94.44 \\
\midrule
\multicolumn{6}{l}{\textit{\textbf{Interpolation without image anchors}}} \\
unbiased interpolation  & I, V & $\times$ & - & 87.38 & 94.57 \\
video-biased  & I, V & $\times$ & $\lambda < 0.5$ & 87.07 & 94.62 \\
image-biased  & I, V & $\times$ & $\lambda > 0.5$ & 87.61 & 94.99 \\
\midrule
\multicolumn{6}{l}{\textit{\textbf{Interpolation with image anchors}}} \\
unbiased interpolation & I, V & $\checkmark$ & - & 87.49 & 94.58 \\
video-biased & I, V & $\checkmark$ & $\lambda < 0.5$ & 87.09 & 94.33 \\
image-biased (\textbf{Ours}) & I, V & $\checkmark$ & $\lambda > 0.5$ & \textbf{88.02} & \textbf{95.10} \\
\bottomrule
\end{tabular}
}
\end{table}

\subsubsection{Components of Feature Alignment}
Table~\ref{table:ablation_alignment} analyzes three key components of the proposed video-oriented feature alignment: 1) \textit{whether interpolation between image and video is necessary}, 2) \textit{whether image features should serve as anchors}, and 3) \textit{how the interpolation should be biased}. 
\textit{Feature Source} specifies the interpolation sets, where \textit{I} and \textit{V} denote the image set $\mathcal{Z}_s$ and video-frame set $\mathcal{Z}_f$, respectively. 
\textit{Image Anchor} indicates whether the anchor is restricted to image features, i.e., $z_a \in \mathcal{Z}_s$ when enabled and $z_a \in \mathcal{Z}_s \cup \mathcal{Z}_f$ otherwise. 
\textit{Bias} denotes the interpolation direction, with $\lambda > 0.5$ moving toward the image feature and $\lambda < 0.5$ toward the video-frame feature.
The first two rows serve as baselines, while the remaining rows evaluate the three design choices.

\textit{Single-domain mixup is inadequate for feature alignment.}
Rows 3-4 show that image mixup provides limited gains, while video mixup underperforms UVIF without alignment. Image mixup regularizes the model but does not reduce the image-video distribution gap, whereas video mixup operates between video frames and may introduce noise to the image head.

\textit{Image anchors facilitate feature alignment.}
Removing the anchor constraint (rows 5-7) allows arbitrary interpolation between image and video features, including between two video features, which degenerates into video mixup and impairs the image head.
In contrast, image anchors (rows 8-10) keep interpolated features between the image and video distributions, preserving compatibility with the image head while enabling adaptation to video representations.

\textit{Image-biased interpolation is critical for feature alignment.}
The comparison among unbiased, video-biased, and image-biased interpolation shows that the first two strategies are suboptimal, while image-biased interpolation achieves the best performance, improving accuracy from 87.30\% to 88.02\%. This suggests that interpolation far from the image distribution can make it difficult for the image head to adapt. In contrast, image-biased interpolation keeps features close to the image anchor while incorporating video information, improving adaptation to video representations.

\begin{figure*}[!t]
    \centering
    \centering
    \begin{minipage}[t]{0.32\textwidth}
        \vspace{0pt}
        \begin{tikzpicture}
            \node[anchor=north west, inner sep=0] (img) at (0,0) {%
                \includegraphics[width=\linewidth]
                {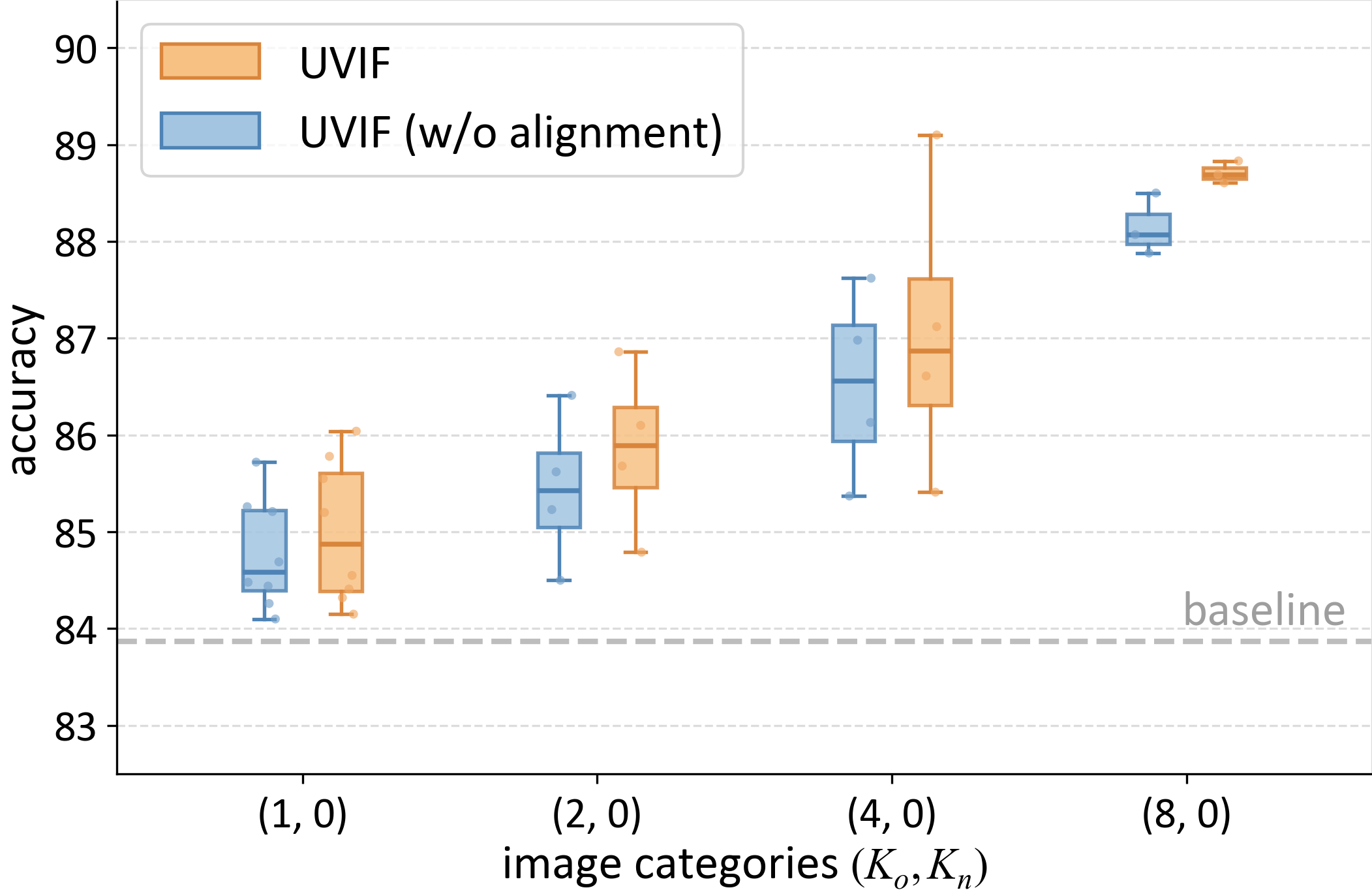}%
            };
            \node[
                anchor=north west,
                inner sep=0,
                xshift=-6pt,
                yshift=-1pt,
                font=\small
            ] at (img.north west) {(a)};
        \end{tikzpicture}
    \end{minipage}
    \hfill
    \begin{minipage}[t]{0.32\textwidth}
        \vspace{0pt}
        \begin{tikzpicture}
            \node[anchor=north west, inner sep=0] (img) at (0,0) {%
                \includegraphics[width=\linewidth]
                {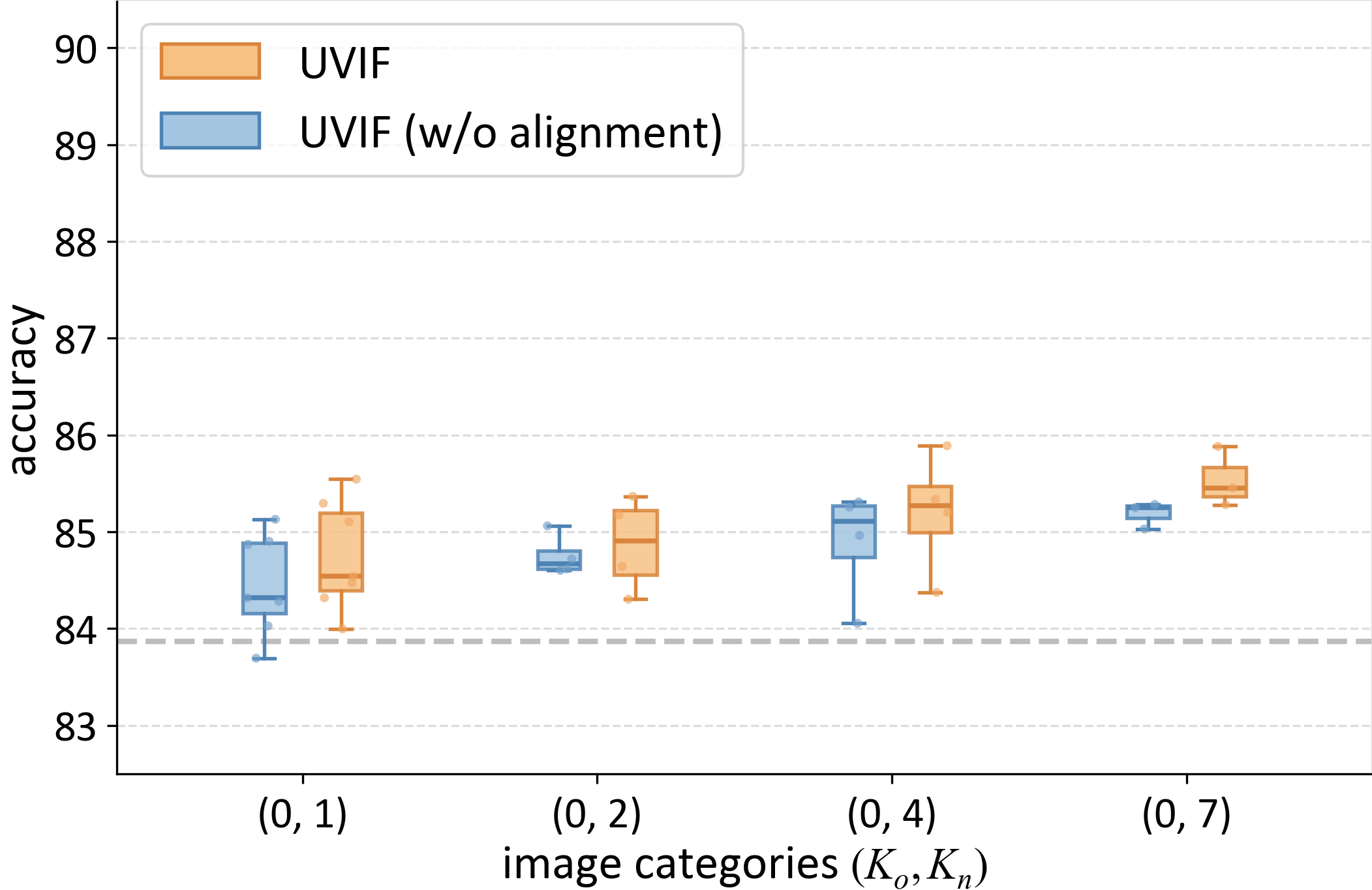}%
            };
            \node[
                anchor=north west,
                inner sep=0,
                xshift=-6pt,
                yshift=-1pt,
                font=\small
            ] at (img.north west) {(b)};
        \end{tikzpicture}
    \end{minipage}
    \hfill
    \begin{minipage}[t]{0.32\textwidth}
        \vspace{0pt}
        \begin{tikzpicture}
            \node[anchor=north west, inner sep=0] (img) at (0,0) {%
                \includegraphics[width=\linewidth]
                {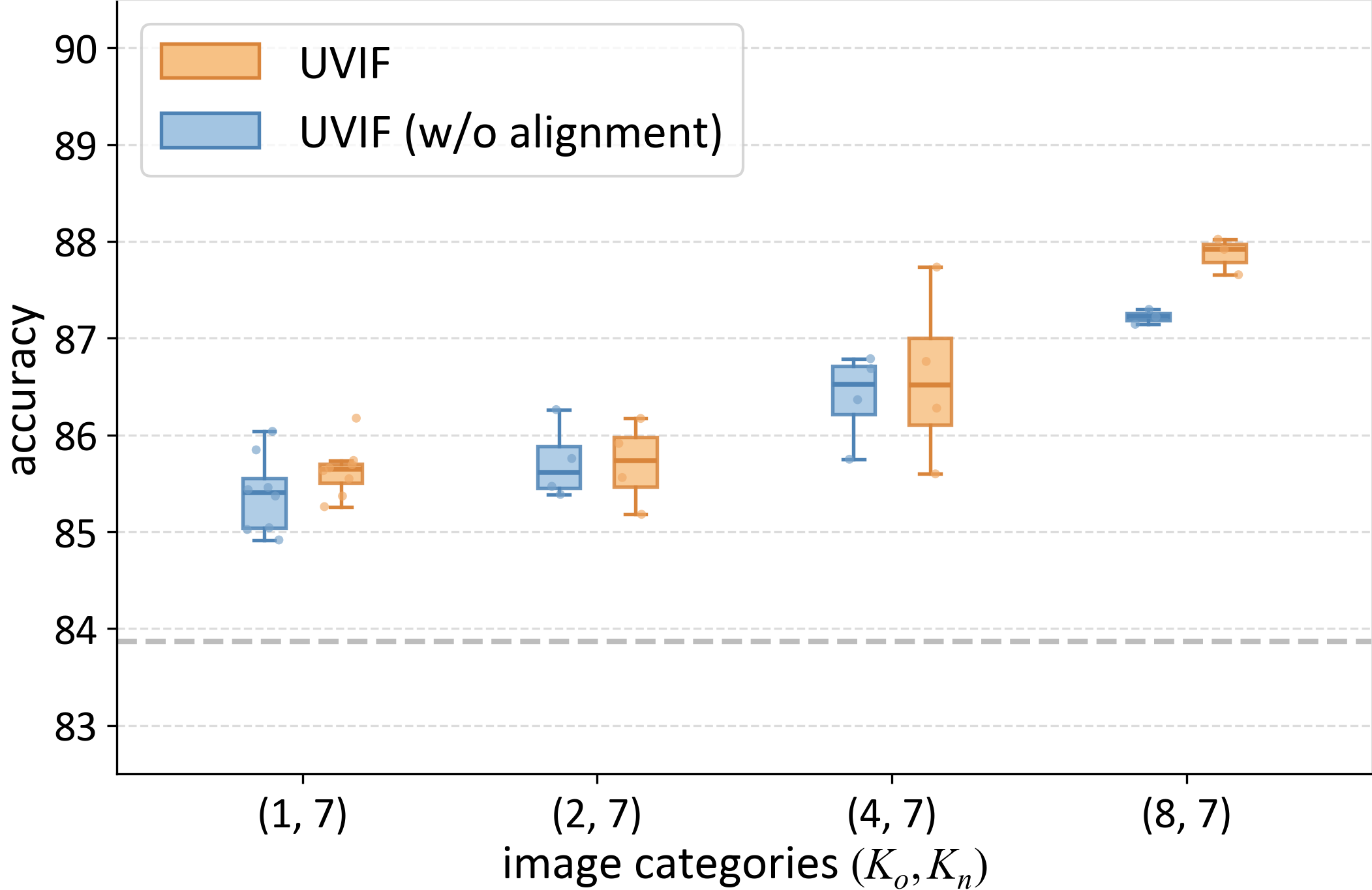}%
            };
            \node[
                anchor=north west,
                inner sep=0,
                xshift=-6pt,
                yshift=-1pt,
                font=\small
            ] at (img.north west) {(c)};
        \end{tikzpicture}
    \end{minipage}
    \caption{Ablation on the compositions of manipulation types in the image training set for UVIF under three settings: (a) Overlap, (b) Non-overlap, and (c) Mixed, denoted by $(K_o, K_n)$. $K_o$ and $K_n$ denote the numbers of overlapping and non-overlapping manipulation types between the video and image sets.}
    \label{figure:ablation_category}
\end{figure*}
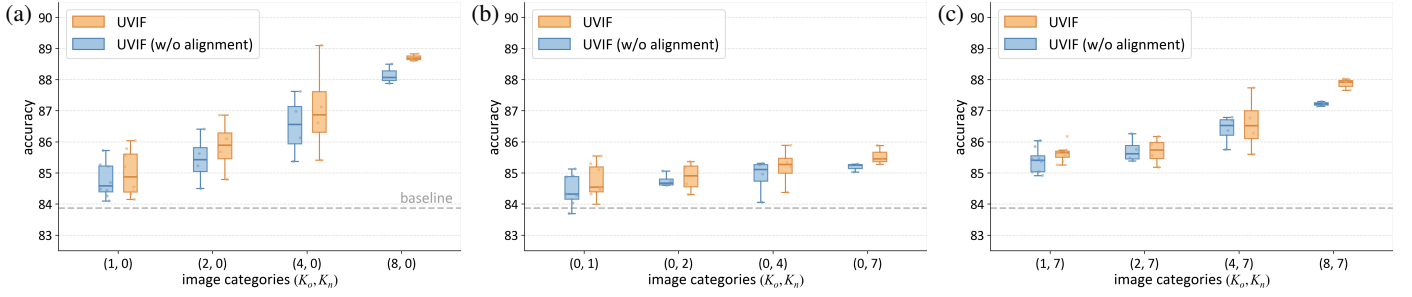

\subsubsection{Composition of Training Images}

We further investigate how the composition of manipulation types in the training image set affects UVIF performance, as shown in Figure \ref{figure:ablation_category}. Specifically, the ForgeryNet video set contains eight manipulation types, while the image set covers these eight overlapping types and seven additional non-overlapping types.
We construct training image sets with different compositions of manipulation types, denoted by $(K_o, K_n)$, where $K_o$ and $K_n$ represent the numbers of overlapping and non-overlapping types, respectively. We consider three groups of settings: 1) \textit{Overlap}, where $K_o\!\in\!\{1,2,4,8\}$ and $K_n\!=\!0$; 2) \textit{Non-overlap}, where $K_o\!=\!0$ and $K_n\!\in\!\{1,2,4,7\}$; and 3) \textit{Mixed}, where $K_o\!=\!8$ and $K_n\!\in\!\{1,2,4,7\}$. For each setting, we randomly sample 100k training samples from the corresponding manipulation types and real images, and summarize the results across different manipulation type combinations.

The results show that the composition of manipulation types in the image training set affects UVIF performance. Under the \textit{Overlap} setting, performance steadily improves as $K_o$ increases from 1 to 8, reaching 88.71\% on average. This indicates that UVIF benefits from greater overlap between the manipulation types in the video and image sets. However, this setting is idealized because the manipulation types in the video set may be unknown.
The \textit{Non-overlap} setting is more challenging because the training video and image sets share no manipulation types. Nevertheless, UVIF consistently improves performance, with further gains as the number of manipulation types increases.
The \textit{Mixed} setting confirms this trend, indicating that more diverse training images help UVIF learn more generalizable representations.
Moreover, UVIF outperforms its variant without feature alignment under all three settings, indicating that feature alignment consistently reduces the image-video distribution gap and improves knowledge transfer from images to videos.

\subsubsection{Different Backbones}
We conduct ablation experiments using different backbones for the unified encoder, as presented in Table \ref{table:ablation_backbones}. 
The evaluated backbones include ResNet \cite{resnet_he2016}, ConvNeXt \cite{convnext_woo2023}, and Swin Transformer \cite{videoswin_liu2022}.
The results show that UVIF consistently improves detection performance across backbone architectures and capacities, demonstrating its generalization across diverse backbones.

\begin{table}[!t]
\centering
\caption{Comparing different backbones for the unified encoder. The baseline is a video classification model.}
\label{table:ablation_backbones}
\begin{tabular}{
    l
    *{2}{>{\centering\arraybackslash}p{0.8cm}}
    *{2}{>{\centering\arraybackslash}p{1.3cm}}
    }
    \toprule
    \multirow{2}{*}{\raisebox{-1ex}{Backbone}}
    & \multicolumn{2}{c}{Baseline}
    & \multicolumn{2}{c}{UVIF (\textbf{Ours})} \\
    \cmidrule(lr){2-3} \cmidrule(lr){4-5}
    & Acc & AUC & Acc & AUC \\
    \midrule
    ConvNeXt-T \cite{convnext_woo2023} 
    & 83.18 & 89.10 
    & 86.32$_{\uparrow\mathrm{3.14}}$ 
    & 94.36$_{\uparrow\mathrm{5.26}}$ \\
    
    R-50 \cite{resnet_he2016} 
    & 83.60 & 90.12 
    & 85.51$_{\uparrow\mathrm{1.91}}$ 
    & 93.12$_{\uparrow\mathrm{3.00}}$ \\
    
    Swin-T \cite{videoswin_liu2022} 
    & 83.89 & 90.53 
    & 88.02$_{\uparrow\mathrm{4.13}}$ 
    & 95.10$_{\uparrow\mathrm{4.57}}$ \\
    
    Swin-S \cite{videoswin_liu2022} 
    & 84.56 & 91.19 
    & 88.94$_{\uparrow\mathrm{4.38}}$ 
    & 95.69$_{\uparrow\mathrm{4.50}}$ \\
    
    Swin-B \cite{videoswin_liu2022} 
    & 84.88 & 90.93 
    & 88.87$_{\uparrow\mathrm{3.99}}$ 
    & 95.84$_{\uparrow\mathrm{4.91}}$ \\
    \bottomrule
\end{tabular}
\vspace{-1em}
\end{table}

\subsubsection{Visualization}
\begin{figure}[t]
    \centering
    \includegraphics[width=\linewidth]{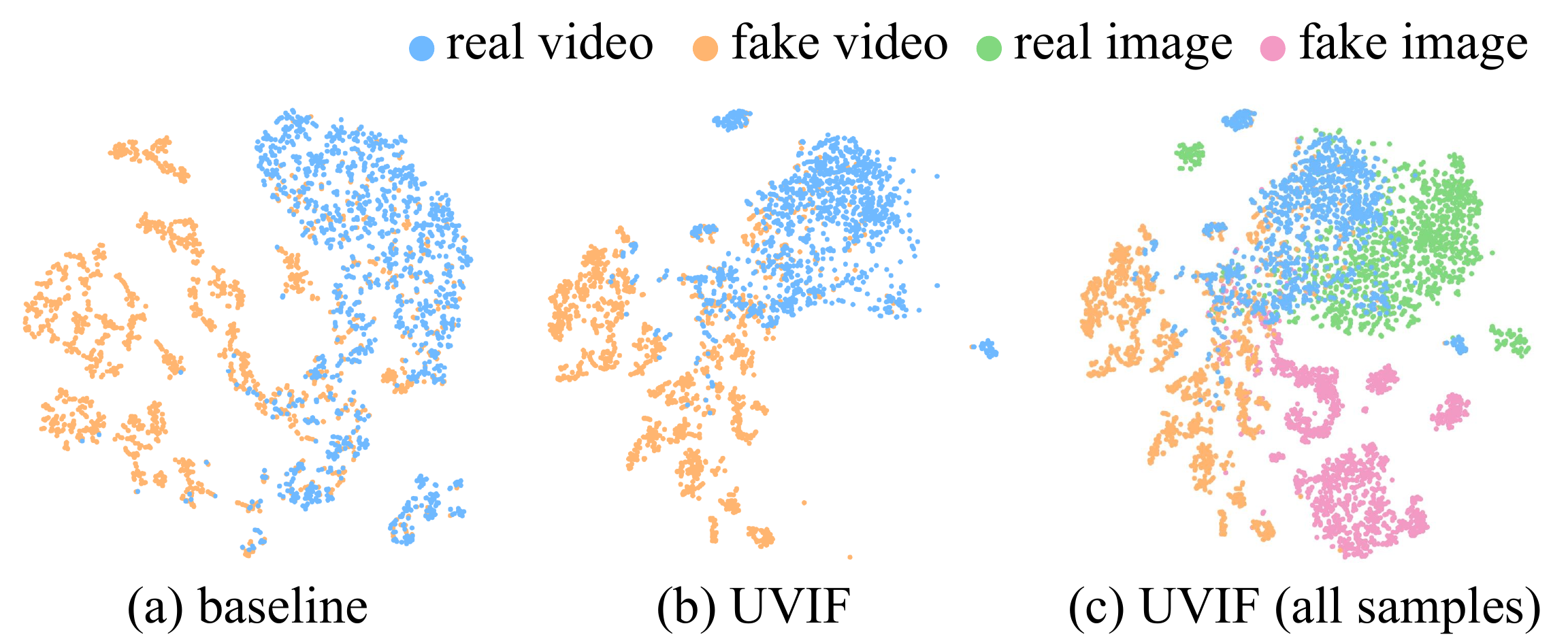}
    \vspace{-2em}
    \caption{T-SNE visualization of the features on ForgeryNet: (a) video classification baseline, (b) UVIF, and (c) UVIF with training images visualized.}
    \label{figure:visual_tsne}
\end{figure}

We further analyze the learned representations through t-SNE visualization, as shown in Figure~\ref{figure:visual_tsne}. We compare the video classification baseline with UVIF on ForgeryNet~\cite{forgerynet_he2021}. The baseline partially separates real and fake video samples, but the two classes still overlap. In contrast, UVIF better separates real and fake samples while forming more compact clusters within each class.
Image and video samples from the same class are also closer in the feature space, indicating that UVIF uses image supervision to learn more discriminative representations for video forgery detection and aligns representations across images and videos.

\section{Conclusion}
In this paper, we present UVIF, an end-to-end multi-task learning framework for video face forgery detection. 
Our method establishes a unified representation of facial videos and images by processing them within a single model.
By utilizing the fine-grained annotations from the image set, UVIF brings significant performance gains for detecting partially forged videos.
In the future, we will investigate face forgery detection from a weakly supervised learning perspective and develop multimodal foundation models for forgery analysis.

\bibliographystyle{IEEEtran}
\bibliography{ref}

 





\end{document}